\documentclass[]{article}

\usepackage{times}
\usepackage{graphicx} 
\usepackage{subfigure} 
\usepackage[round]{natbib}
\usepackage{algorithm}
\usepackage{algorithmic}
\usepackage{amsthm}

\usepackage{amsmath}
\usepackage{amsfonts}
\usepackage{bbm}
\usepackage{url}
\usepackage{hyphenat}
\usepackage{amssymb}
\usepackage{multicol}

\usepackage[margin=1.0in]{geometry}

\newcommand{\covariateSpace}{\mathbb{R}^D}
\newcommand{\nTrainingPoints}{N}
\newcommand{\labelSpace}{\mathcal{L}}
\newcommand{\pointProcess}{X}
\newcommand{\sample}{\mathbf{x}}
\newcommand{\point}{x}
\newcommand{\intensityFunction}{\rho}
\newcommand{\testSet}{A}
\newcommand{\nTestSet}{n(\testSet)}
\newcommand{\intensityMeasure}{\eta}
\newcommand{\meanFunction}{\mu}
\newcommand{\covarianceFunction}{C}
\newcommand{\productDensity}{m}
\newcommand{\nClasses}{Q}
\newcommand{\observationSample}{\mathbf{y}}
\newcommand{\superpositionProductDensity}{M}
\newcommand{\softMax}{\mathcal{S}}
\newcommand{\numberP}{\# \mathcal{P}}
\newcommand{\nProduct}{K}
\newcommand{\categoryIndex}{i}
\newcommand{\pointIndex}{j}
\newcommand{\pointIndexB}{k}
\newcommand{\janossyDensity}{\theta}
\newcommand{\kernelDensity}{\beta}
\newcommand{\kernelDensityFunction}{G}

\newcommand{\observation}{y}
\newcommand{\densityArgument}{(x_1  , x_2, \hspace{2 pt} \ldots \hspace{2 pt} , x_\nProduct )}
\newcommand{\localDensityArgument}{(x_1  , x_2, \hspace{2 pt} \ldots \hspace{2 pt} , x_\nProduct | A )}
\newcommand{\densitySpace}{(\mathbb{R}^D) ^{ \hspace{1 pt } \nProduct}}
\newcommand{\thinning}{\tau}
\newcommand{\thinningParameter}{\gamma}

\newtheorem*{kAndZ}{Theorem 1}
\newtheorem*{mAndG}{Theorem 2}

\title{Classification using log Gaussian Cox processes}

\author{ {\bf Alexander G. de. G Matthews } \\ \vspace{-10pt}  \\ University of Cambridge \\ \vspace{-12pt} \\ am554@cam.ac.uk \and {\bf Zoubin Ghahramani} \\ \vspace{-10pt} \\ University of Cambridge  \\ \vspace{-12pt} \\ zoubin@eng.cam.ac.uk }

\date{}

\begin{document} 

\maketitle

\begin{abstract} 
McCullagh and Yang \citeyearpar{McCullagh2006} suggest a family of classification algorithms based on Cox processes. We further investigate the log Gaussian variant which has a number of appealing properties. Conditioned on the covariates, the distribution over labels is given by a type of conditional Markov random field. In the supervised case, computation of the predictive probability of a single test point scales linearly with the number of training points and the multiclass generalization is straightforward. We show new links between the supervised method and classical nonparametric methods. We give a detailed analysis of the pairwise graph representable Markov random field, which we use to extend the model to semi-supervised learning problems, and propose an inference method based on graph min-cuts. We give the first experimental analysis on supervised and semi-supervised datasets and show good empirical performance.
\end{abstract} 

\section{Introduction}

Classification is a central problem in statistics and machine learning. Despite the simplicity of the problem specification, and a literature with an age comparable to the fields themselves, it is still an active and diverse research area. This can be explained in part because of the vast variety of applications, which only grows with time.

In this paper we analyze a log Gaussian Cox process (LGCP) model for classification. This model was originally proposed by McCullagh and Yang \citeyearpar{McCullagh2006} as a part of a family of Cox process based models. It has received comparatively little attention in the machine learning and artificial intelligence communities despite being of particular interest when scalable, tractable inference is important.

Perhaps surprisingly, conditioned on the sample covariates, the model reduces to a Potts distribution \citep{Wu1982,McCullagh2008} for the labels, with the energies simply related to the mean and covariance function of the Gaussian process. There are thus strong connections to familiar models such as the Boltzmann machine \citep{Ackley1985} and the model can be viewed as a prescribed case of a conditional random field \citep{Lafferty2001}. It is possible to leverage the wealth of literature on inference in such models for both supervised and semi-supervised learning in the LGCP, as we shall show. 

The model may be interpreted as a Bayesian nonparametric one \citep{Orbanz2010}. Specifying a model in such a way is desirable because it avoids overfitting (i.e Bayesian) and has large support (i.e nonparametric). The particular model in this case is a marked Cox process \citep{Cox1955}, which is a particular type of spatial point process \citep{Daley2003}. The latent intensity for each class is modelled using an exponentiated Gaussian process \citep{Moller1998}.  

In the supervised case there are intriguing relations to classical nonparametric methods such as kernel density estimation \citep{Rosenblatt1956,Parzen1962} and nearest neighbour classifiers, raising interesting questions about the interplay between classical nonparametric statistics and Bayesian nonparametric methods.

It is desirable that inference in a model be simple and robust. Scaling to large datasets can often be challenging with Bayesian nonparametric methods. For example, the intrinsic computational complexity of Gaussian process regression is $O(\nTrainingPoints^3)$ \citep{Rasmussen2005}  where $\nTrainingPoints$ is the number of training points. Further, Gaussian process classification requires approximate inference. By contrast, the supervised LGCP model has $O(\nTrainingPoints)$ closed form computation for prediction on a test point which means that we have been able to investigate our model on some large datasets. So far fairly simple parameter training using cross validation has been sufficient, though it seems likely this could be developed using the literature on Markov random fields \citep{Koller2009}.

\subsection{Relation to existing work and contributions of this paper}

As we have already described, this work builds on the excellent contribution of McCullagh and Yang \citeyearpar{McCullagh2006}, which has been discussed in subsequent literature.

In the first paper of McCullagh and Yang \citeyearpar{McCullagh2006}, the LGCP classifier and the related permanental Cox process model are first considered. The supervised model equations, including the predictive equations (\ref{eq:predictiveDistribution}) are motivated for the first time but the LGCP model is not implemented or tested empirically. A connection to Gibbs partitions \citep{Pitman2006} is mentioned. 

In a paper on bias in logistic models \citep{McCullagh2008}, the two models are used as an illustrative example of a non-standard supervised classification model. The relation between the Janossy density and product density is discussed in the context of different sampling schemes and a new derivation (which we have followed) is given. The connections to Gaussian process classification, here seen as a special case of logistic regression with random effects, are discussed. There is a single sentence describing the connection to Markov random fields \citep{Besag1974}.

In a paper on the permanental Cox process classifier \citep{McCullagh2012}, which could be viewed as an analogue of this paper, the permanental variant is implemented using an approximation scheme based on a cyclic approximation, and then tested on real data. For further discussion of the relation between the permanental model and the log Gaussian model see \citep{McCullagh2006,McCullagh2008} and for some summary comments given for the perspective of machine learning see the appendix.

In the field of semi-supervised learning, Markov random fields and the use of graph min-cuts for MAP inference have been considered before in the literature \citep{Zhang2001,Blum2001} but not with a Bayesian nonparametric motivation. As shall be seen in section \ref{section:SSL} the Bayesian nonparametric model places specific requirements on the edge weights of the Markov random field that are prescribed by the prior. \\

\noindent To summarize, the contributions of our work are, in broad terms:

\begin{itemize}
\item The recognition of the LGCP classification model from McCullagh and Yang \citeyearpar{McCullagh2006}, which despite a variety of appealing characteristics has not, to our knowledge, been previously used in practice. \item The discovery of new connections between the supervised LGCP method and classical nonparametric methods as discussed in section \ref{section:classical}. These additional connections prove important in explaining the experimental results. 
\item The first empirical validation of the supervised method against some commonly used datasets in section \ref{section:supervisedResults}, including some large, high dimensional datasets. Large dataset results are of particular interest in Bayesian nonparametrics, where they can be harder to obtain.
\item To elaborate, in section \ref{section:MRF}, the general form of the crucial link to Markov random fields, after the initial sentence from McCullagh \citeyearpar{McCullagh2008}. 
\item To show, in section \ref{section:submodularity}, that the Markov random field is pairwise graph-representable in the sense of Kolmogorov and Zabih \citeyearpar{Kolmogorov2004} and hence pairwise submodular.
\item To extend the model to semi-supervised learning in section \ref{section:SSL}.
\item To propose robust inference for the semi-supervised learning model using graph min-cut algorithms, that exploit the pairwise graph-representable property, as described in section \ref{section:SSL}. 
\item The experimental validation of the proposed semi-supervised learning algorithm on real data as described in section \ref{section:sslExperiments}.
\end{itemize}

In this paper we now proceed by providing a summary of the necessary background theory, which we have strived to make relatively self-contained.

\section{Background}

We will be interested in this paper in spatial point processes \citep{Daley2003,Moller2004}. The realisations of spatial point processes are sets $\pointProcess$ of points $\point$ in some space. In this paper all the points will lie in $\covariateSpace$.  Each point will be supplemented with a mark, which in this case will be in some finite label set $\labelSpace$. The pairs of points and labels constitute what is called a marked point process.

The well known Poisson process \citep{Kingman1993} is a simple spatial point process that can be used to build more complex models. It is parameterized by a non-random intensity function $ \intensityFunction : \covariateSpace \mapsto [0,\infty) $ which has the property of being locally integrable, so that $ \int_{\testSet} \intensityFunction(\point) \mathrm{d} \point < \infty $ for all bounded Borel sets $\testSet \subset \covariateSpace$. This can be used to define an intensity measure so that 

\begin{equation}
\intensityMeasure(\testSet) = \int_{\testSet} \intensityFunction( \point ) \mathrm{d} \point
\end{equation}

This measure is assumed to be diffuse, which means that $\intensityMeasure(\lbrace \point \rbrace) = 0  \hspace{5 pt} \forall \point \in \covariateSpace $. The Poisson process can then be defined in terms of the distribution over the number of points in any given Borel set $A$, which we denote as $n(A)$. 

\begin{enumerate}
\item For Borel set $\testSet \subset \covariateSpace : \intensityMeasure(\testSet) < \infty $, the distribution of $\nTestSet$ is Poisson, with rate parameter $\intensityMeasure(\testSet)$ if $ \intensityMeasure(\testSet) \neq 0 $ and $\nTestSet = 0$ otherwise.
\item For Borel sets $\testSet, B$ such that $\testSet \cap B = \emptyset $, $\nTestSet$ is independent of $n(B)$.
\end{enumerate}

The Cox process \citep{Cox1955} can be defined in terms of the Poisson process. Instead of a non-random intensity $\intensityFunction$, we now make it a well-behaved non-negative random function. Then conditioned on this random function, the points from a Cox process are distributed as a Poisson process with that intensity.

The case where the random function is given by the exponential of a Gaussian process is called a log Gaussian Cox process \citep{Moller1998}.

\begin{align}
f &\sim \mathcal{GP}(\meanFunction,\covarianceFunction) \\
\intensityFunction &= \exp(f)
\end{align}

Here $\mathcal{GP}$ denotes a Gaussian process \citep{Rasmussen2005}. Following the exposition of M\o ller et al \citeyearpar{Moller1998}, some restrictions on the form of this Gaussian process are assumed in this paper. First it is assumed that the process is stationary, which will mean that the mean function $\meanFunction$ is constant and that the covariance function $\covarianceFunction(x_\pointIndex,x_\pointIndexB)$ has the form of a function $\covarianceFunction(x_\pointIndex-x_\pointIndexB)$. Further it is assumed that there exist $\alpha, \beta > 0 $ such that:

\begin{equation}
1 - \frac{\covarianceFunction(s)}{\covarianceFunction(0)} < \alpha ||s||^{\beta} \hspace{4 pt} \forall s \hspace{2pt} :  \hspace{2pt} ||s||<1
\end{equation}

This last condition ensures the intensity measure $\intensityMeasure$ is almost surely a continuous modification of $f$. These are sufficient, but not necessary conditions for a well defined LGCP. For further detail on such matters see the paper of M\o ller et al \citeyearpar{Moller1998}. Examples of valid kernels include the commonly used squared exponential, and exponential kernels.

\begin{figure}[ht!]
\centerline{\includegraphics[width=3.0in]{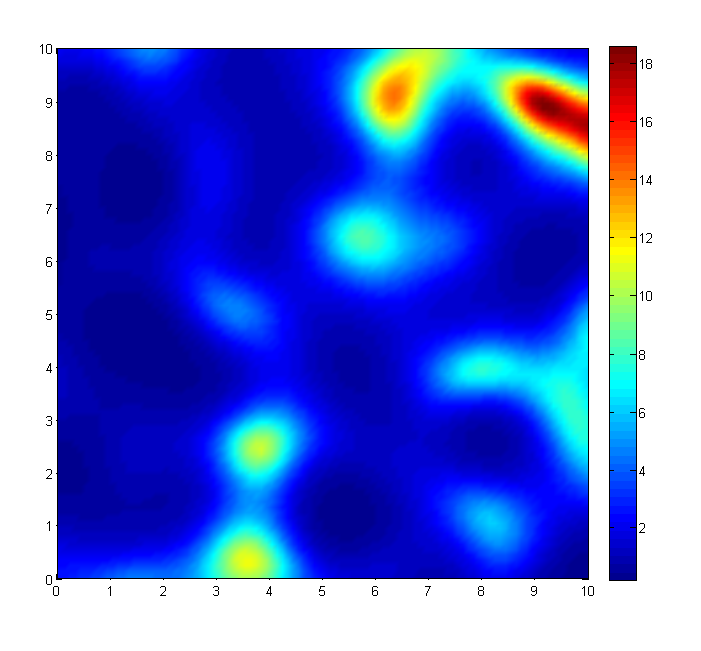} \includegraphics[width=3.0in]{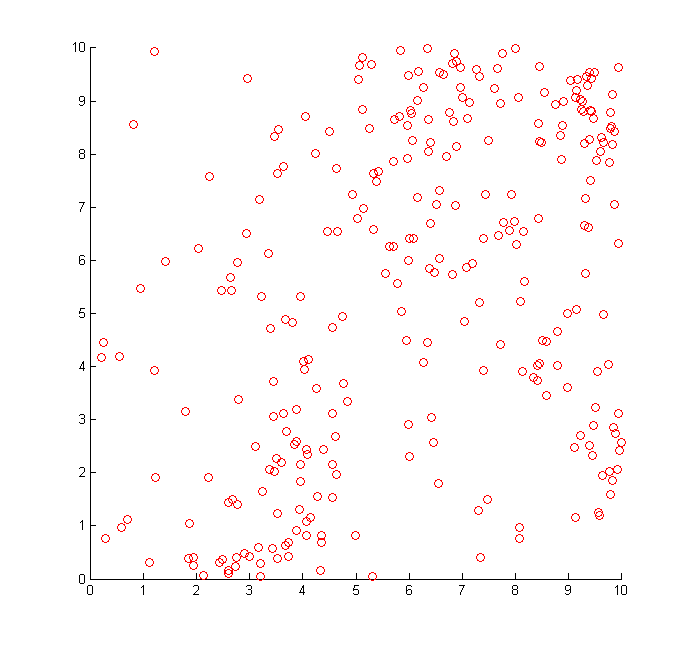}}
\caption{ \label{fig:LGCPSamples}  Left: sample of random intensity function for a log Gaussian Cox process with isotropic squared exponential kernel with unit variance and length scale.
Right: sample points drawn with a Poisson rate from the same intensity function (see text)}
\end{figure}

Figure (\ref{fig:LGCPSamples}) demonstrates the LGCP in two dimensions. The Gaussian process had mean zero and a squared exponential kernel, with unit length scale and signal variance. 
A realisation of the random intensity of the Cox process and corresponding sample points are shown.  

An important set of quantities for describing a Cox process, if they exist is the product densities \citep{Daley2003}. The $\nProduct$-th product density, where $K$ is a positive integer, is a function $\productDensity_{[\nProduct]}\densityArgument $ of points $x_\pointIndex \in \covariateSpace $, for each $ \pointIndex = 1,...,\nProduct $. An intuitive interpretation of the product density can be given via the equation:

\begin{equation}\label{eq:productDensityInterpretation}
\begin{aligned}
& \productDensity_{[\nProduct]}\densityArgument dx_1 dx_2 ... dx_\nProduct \\
& = Pr( \mbox{ One point in each small volume \hspace{1 pt}}(x_\pointIndex, x_\pointIndex + dx_\pointIndex ) )
\end{aligned}
\end{equation}

This can be contrasted to the $\nProduct$-th local Janossy density $ \janossyDensity_\nProduct\localDensityArgument $,  defined relative to some bounded Borel set $A \subset \covariateSpace$ which, when it exists, has the property:

\begin{equation}
\begin{aligned}
&\janossyDensity_\nProduct\localDensityArgument dx_1 dx_2 ... dx_\nProduct \\
=Pr( \mbox{ Exactly } \nProduct \mbox{ points}& \mbox{ in A, one in each small volume \hspace{1 pt}} (x_\pointIndex,x_\pointIndex+dx_\pointIndex) \mbox{ and none elsewhere in A} )
\end{aligned}
\end{equation}

Clearly, the two differ in the sense that the local Janossy density quantifies occurrence of points and non-occurrence of points elsewhere, whereas the product density only quantifies the occurrence of points in a certain region, and does not rule out the possibility of points elsewhere. The local Janossy density is the density of the local Janossy measure and the product density is the density of the factorial moment measure, both with respect to Lebesgue measure on $\densitySpace$. For more details see Daley and Vere-Jones \citeyearpar{Daley2003}.

In the case of the LGCP, subject to conditions that are always met in this paper, the product densities exist \citep{Moller1998} and are given by:

\begin{equation} \label{eq:LGCPproductDensity}
\productDensity_{[\nProduct]}\densityArgument = \exp \left \lbrace \nProduct \meanFunction + \frac{1}{2} \sum_{\pointIndex,\pointIndexB = 1}^{\nProduct} \covarianceFunction(x_\pointIndex-x_\pointIndexB) \right \rbrace
\end{equation}

For what follows we shall, for convenience, refer to the product densities and Janossy densities in the singular since the dependence on $\nProduct$ is obvious from the context.

\section{Model}

\subsection{Model definition}

In this section we review the model for LGCP classification given in the original paper by McCullagh and Yang \citeyearpar{McCullagh2006}, although the treatment of the observation model follows the later work of McCullagh \citeyearpar{McCullagh2008}.  

A fixed, known, number of classes $\nClasses$ is assumed. Each class population $\categoryIndex=1,...,\nClasses$ is modelled using a separate independent log Gaussian Cox process $\pointProcess_\categoryIndex$ with mean function $\meanFunction^{(\categoryIndex)}$ and covariance function $\covarianceFunction^{(\categoryIndex)}$ . 

\begin{align}
f_\categoryIndex \sim \mathcal{GP}&(\meanFunction^{(\categoryIndex)},\covarianceFunction^{(\categoryIndex)}) \hspace{ 5 pt} \categoryIndex=1,...,\nClasses \\ 
\intensityFunction_\categoryIndex &= \exp(f_\categoryIndex) \\
X_\categoryIndex \hspace{2 pt} | \hspace{2 pt} \intensityFunction_\categoryIndex &\sim \mathcal{PP}(\intensityFunction_\categoryIndex)
\end{align}
Here $\mathcal{PP}(\cdot)$ denotes a Poisson process with a given intensity. If one considers the superposition of the $\nClasses$ different Cox processes and ignores the labels, another Cox process $\pointProcess$ is obtained, which is termed the superposition process.
\begin{align}
X &= \cup_{\categoryIndex=1}^{\nClasses}X_\categoryIndex
\end{align}

This Cox process may be associated with the total intensity $\intensityFunction_T = \sum_{\categoryIndex = 1}^{Q} \intensityFunction_\categoryIndex$. The superposition Cox process is not a \emph{log Gaussian} Cox process since the distribution for $\intensityFunction_T$ cannot be found by exponentiating any Gaussian process.

Next a method is defined by which samples from the population come to be observed. Let $\thinning(S,\thinningParameter)$ be the spatial point process whose points lie in $\mathbb{R}^D$ resulting from an independent random thinning of the points in $S$. Loosely speaking this means that with probability $\thinningParameter$ a point in $S$ is present in the new set of points. It is assumed that this thinning procedure is applied to each of the class populations to obtain new thinned category samples $T_\categoryIndex$:

\begin{equation}
T_\categoryIndex \hspace{2 pt} | \hspace{2 pt} X_\categoryIndex \sim \thinning( X_\categoryIndex, \thinningParameter )
\end{equation}

\vskip 0.1in 
One may also define the thinned superposition $ T = \cup_{\categoryIndex=1}^{\nClasses}T_\categoryIndex $. Consider fixing $A$ to be the bounded Borel set in which observations are performed and let $ \sample = A \cap T $ be the observed covariates. 
Now define $\sample^{(\categoryIndex)} = \lbrace \point_\pointIndex \in \sample : y_\pointIndex = \categoryIndex \rbrace $, which denotes the set of points in the sample that have label $\categoryIndex$. The conditional probability, for a given value of $\thinningParameter$ is equal to:

\begin{equation}
Pr_\thinningParameter( \observationSample | \sample ) = \frac{ \mathbb{E}\left[  \exp\lbrace - \thinningParameter \int_A \intensityFunction_T(x') dx' \rbrace \prod_{\pointIndex=1}^{N} \intensityFunction_{y_\pointIndex}(x_\pointIndex) \right]_{\lbrace \intensityFunction_\categoryIndex \rbrace}}{\mathbb{E}\left[ \exp\lbrace - \thinningParameter \int_A \intensityFunction_T(x') dx' \rbrace \prod_{\pointIndex=1}^{N} \intensityFunction_T(x_\pointIndex) \right]_{\lbrace \intensityFunction_\categoryIndex \rbrace}}
\end{equation}

\vskip 0.05in
\noindent where the expectation is taken over the set of intensity functions $\lbrace \intensityFunction_\categoryIndex \rbrace_{\categoryIndex = 1}^{\nClasses}. $ Consider the limit $\thinningParameter \rightarrow 0 $ which represents an idealized observation process. In the limit the distribution of the labels is given by:

\begin{equation}\label{eq:conditionalModel}
Pr( \observationSample ) [\sample ] = \lim_{\thinningParameter \rightarrow 0 } \left [ Pr_\thinningParameter( \observationSample | \sample  ) \right] = \frac{\prod_{\categoryIndex=1}^{\nClasses}\productDensity^{(\categoryIndex)}(\sample^{(\categoryIndex)})}{M(\sample)} 
\end{equation}

$ \productDensity^{(\categoryIndex)}(\cdot) $ is the product density of Cox process $\categoryIndex$ and is given by equation (\ref{eq:LGCPproductDensity}) evaluated with the corresponding mean function $\meanFunction^{(\categoryIndex)}$ and covariance function $\covarianceFunction^{(\categoryIndex)}$. $\superpositionProductDensity(\cdot)$ is the product density of the superposition process, which, as we have discussed, does not have the form of the product density for an LGCP. We will investigate its form in the sections that follow. It would be tempting to denote this distribution using the conditional notation `$Pr( \observationSample |\sample )$' but we use the non-standard notation $Pr( \observationSample ) [\sample]$ to highlight the limiting construction being used.

Post-hoc it can be seen that the label distribution in equation (\ref{eq:conditionalModel}) has an intuitive interpretation in the sense that it is proportional to a product of product densities, one per class divided by the superposition product density. Using the interpretation given in equation (\ref{eq:productDensityInterpretation}) one sees the relationship to the probability that each independent process contains a certain set of points in a small interval around $\sample^{(\categoryIndex)}$ in precisely the sense of a product density as opposed to a Janossy density. Note also that the dependence on the observation Borel set $A$ has vanished in the limit. An advantage of this observation model, which at least partly motivates the approach \citep{McCullagh2008} is the relative tractability of the result.

Substituting the product density from equation (\ref{eq:LGCPproductDensity}) into equation (\ref{eq:conditionalModel}) one obtains:

\begin{align}\label{eq:complexMRF}
Pr( \observationSample ) [\sample] = \frac{1}{M(\sample)}\exp \left\lbrace \sum_{\categoryIndex=1}^{Q}|\sample^{(\categoryIndex)}| \meanFunction^{(\categoryIndex)} + \frac{1}{2} \sum_{\categoryIndex=1}^{Q} \sum_{(x_\pointIndex,x_\pointIndexB) \in \sample^{(\categoryIndex)} \times \sample^{(\categoryIndex)}  } \covarianceFunction^{(\categoryIndex)}(x_\pointIndex-x_\pointIndexB) \right\rbrace 
\end{align}

\noindent where $|\sample^{(\categoryIndex)}|$ denotes the cardinality of the set $\sample^{(\categoryIndex)}$. 

Cox \citeyearpar{Cox1958} and McCullagh \citeyearpar{McCullagh2008} consider the \emph{no interference condition}. In general a discriminative model fulfils the no-interference condition if it has the following property:

\begin{equation}
\sum_{\observationSample^*} Pr( \observationSample^{*}, \observationSample | \sample^{*}, \sample ) = Pr( \observationSample | \sample) 
\end{equation}
 
For example, the multivariate Gaussian marginalization property guarantees that this criterion is fulfilled in the case of Gaussian process classification \citep{Rasmussen2005}. This means that knowledge of the existence of other unlabelled covariates does not affect our predictions about data. McCullagh \citeyearpar{McCullagh2008} discusses the extent to which this is a desirable characteristic of a model. Here we point out that for the concept of semi-supervised learning to make sense in a model the no interference condition must not be fulfilled. Treating the limiting process as standard conditioning the LGCP model does not fulfil the no-interference condition and thus has a meaningful extension to semi-supervised learning as we discuss in section \ref{section:SSL}.

\subsection{Supervised learning}\label{section:supervisedLearning}

The supervised learning case was originally discussed by McCullagh and Yang \citeyearpar{McCullagh2006}. Here we review the existing material and discuss in section \ref{section:classical} some novel connections to classical nonparametric methods. These new connections prove important for understanding the behaviour of the supervised method which will be evaluated empirically for the first time in the experiments section. 

Consider the case where a set of covariates $\sample$ and their labels $\observationSample$ have been observed and one wishes to predict the label $y^{*}$ of a test point with measured covariates $x^{*}$. The requisite distribution is:

\begin{equation} \label{eq:predictiveDistribution}
Pr( y^{*} = \categoryIndex | \observationSample )[\sample \cup \lbrace   x^{*} \rbrace] = \softMax ( F )_\categoryIndex
\end{equation}

We have defined the softmax function $ \softMax \in \mathcal{R}^\nClasses \ $ which has components:

\begin{equation}
\softMax(x)_\categoryIndex = \exp(x_\categoryIndex) / ( \sum_{\beta =1}^{\nClasses} \exp(x_\beta) )
\end{equation}

and $F \in \mathbb{R}^\nClasses$ whose components are given by: 

\begin{equation} \label{eq:activationFunction}
F_\categoryIndex =  \meanFunction^{(\categoryIndex)}  + \frac{1}{2} \covarianceFunction^{(i)}(x^{*},x^{*} ) + \sum_{x \in \mathbf{x}^{(\categoryIndex)} } \covarianceFunction^{(\categoryIndex)}(x^{*}, x )
\end{equation}

In the case where the mean and covariance are the same for each category the first two terms in $F_\categoryIndex$ will cancel when one takes the softmax. Clearly computation of the predictive density is linear in the number of training data points.

\subsubsection{Connections to classical nonparametric methods.}{\label{section:classical}}

We now show a novel relation of this model to \emph{kernel density} or \emph{Parzen\hyp{}Rosenblatt  window } \citep{Rosenblatt1956,Parzen1962} estimates of the individual class densities. These new connections prove important for understanding the behaviour of the supervised method which will be evaluated empirically for the first time in the experiments section.

Given a symmetric non-negative normalized function of two points $\kernelDensityFunction(x,x')$ the simplest kernel density estimate $\kernelDensity(x^{*})$  at a test point $x^*$ given a set of $N$ points $x_\pointIndex$ is given by:

\begin{equation}
\kernelDensity(x^{*}) = \frac{1}{N} \sum_{\pointIndex = 1}^{N} \kernelDensityFunction(x^{*},x_\pointIndex) 
\end{equation}

Now let us assume that we take the kernel density estimate of each class $\categoryIndex$ each containing $N_\categoryIndex$ points and giving $N_T$ points in total. We wish to produce a probabilistic estimate of the class of $y^*$. Allowing for the empirical counts in each class with a factor $\frac{N_\categoryIndex}{N_T}$ we obtain:

\begin{equation}\label{eq:kdePredictive}
Pr(y* = \categoryIndex ) = \frac{\kernelDensity(x^{*})_\categoryIndex N_\categoryIndex}{\sum_{j=1}^{\nClasses} \kernelDensity(x^{*})_j N_j}
\end{equation}

If we stipulate that:

\begin{equation}
\kernelDensityFunction(x,x') = \covarianceFunction^{(\categoryIndex)}(x,x') = \covarianceFunction(x,x') \hspace{5 pt} \forall \categoryIndex  
\end{equation}

\noindent and that $\meanFunction^{(\categoryIndex)}=0 \hspace{3 pt } \forall \categoryIndex $ then the predictive equations 
(\ref{eq:predictiveDistribution}) and (\ref{eq:kdePredictive}) imply some functional similarities and differences between the two models. Clearly the predictive probability distributions are not the same since the LGCP has a softmax function. However under a $0$-$1$ loss the optimal decision rule for classification is to choose the maximum assignment. Both predictive distributions in fact have the same maximum leading to the same decisions. This equivalence extends to the parameter estimation if we use cross-validation on the $0$-$1$ loss but not if we for instance use the negative log of the predictive probability. In the case where the covariance functions are different it would seem hard to `derive' the term $\frac{1}{2} \covarianceFunction^{(i)}(x^{*},x^{*} )$ in equation (\ref{eq:activationFunction}) in terms of the kernel density estimator. 

The kernel density estimator is well studied in nonparametric statistics. In the regimes where they are equivalent, frequentist guarantees derived for the kernel density style algorithms apply equally to LGCP supervised learning. In the regimes where they are not equivalent this would seem like a good starting point to derive such guarantees.

\subsection{Relation to conditional Markov random fields}\label{section:MRF}

Equation (\ref{eq:conditionalModel}) is related to the many types of Markov random field \citep{Besag1974}, as is briefly mentioned in a sentence in a paper of McCullagh \citeyearpar{McCullagh2008}. In this section then sections \ref{section:SSL} and \ref{section:submodularity} we further analyze this link, giving the first detailed exposition. Our augmented characterization will prove essential for the generalization to semi-supervised learning given in section \ref{section:SSL}.

Equation ({\ref{eq:complexMRF}) may be rearranged to give: 

\begin{align}\label{eq:explicitMRF}
Pr( \observationSample ) [\sample]  = \frac{1}{M(\sample)}\exp \left \lbrace \sum_{\pointIndex=1}^{N} \meanFunction^{ \left( y_\pointIndex \right )} + \frac{1}{2} \sum_{\pointIndex,\pointIndexB=1}^{N} \covarianceFunction^{ \left( y_\pointIndex \right )}(x_\pointIndex-x_\pointIndexB) \delta( y_\pointIndex, y_\pointIndexB) \right \rbrace 
\end{align}

\noindent where $\delta$ denotes the Kronecker delta function. The connection to Markov random fields has thus been made explicit. This probability distribution is a type of fully connected Potts model \citep{Wu1982}. In general the pairwise energies depend not only on the covariates $ \lbrace x_\pointIndex \rbrace $ but also on the classes $ \lbrace y_\pointIndex \rbrace$. The product density of the superposition process $M(\sample)$, can be seen to correspond to the partition function or normalizing constant of the Potts model.

The link to the Potts model here differs from their use in some other pattern recognition problems, in the sense that the weights are stipulated by the kernel function of the Gaussian process, which are in turn derived from the observed covariates. In this sense, conditioned on $\sample$, one can also view the resulting Potts model as a kernelized conditional random field \citep{Lafferty2001}.

If we choose to make the simplifying assumption that the means $\meanFunction^{(\categoryIndex)} $ are zero and the kernel functions are all the same $\covarianceFunction^{(\categoryIndex)} = \covarianceFunction$. The conditional model can be written as:  

\begin{equation}\label{eq:simpleMRF}
Pr( \observationSample ) [\sample]  = \frac{1}{M(\sample)}\exp \left \lbrace \frac{1}{2} \sum_{\pointIndex,\pointIndexB=1}^{N} \covarianceFunction(x_\pointIndex-x_\pointIndexB) \delta( y_\pointIndex, y_\pointIndexB) \right \rbrace
\end{equation}

If we take another special case, with two categories ($Q=2$), then we obtain a fully connected Ising model or Boltzmann machine \citep{Ackley1985}. 

In general, both the Ising model and the Potts model partition functions are $\numberP$-Hard to compute, even with the constraint of positive weights. Thus, so is the product density of the superposition process. However, in the positive weight, binary case, it is possible to approximate these partition functions with high probability, to within a given percentage, in polynomial time using Markov chain Monte Carlo methods \citep{Jerrum1993}. 

We delay our discussion of the pairwise graph-representability and submodularity of the Markov random field until section \ref{section:submodularity}.

\subsection{Semi-supervised learning}\label{section:SSL}

In this section, using the connection to the Potts model, we show how to extend the model to semi-supervised learning for the first time, then discuss the relationship to ideas in the existing semi-supervised learning literature.

Consider a partitioning of the data into two sets; the `starred' set $\sample^{*}, \observationSample^{*} $ which will correspond to the unlabelled test data and the `unstarred set' $\sample, \observationSample $ which will correspond to the labelled data. Conditioned on the training covariates $\sample$ and a set of test covariates $\sample^{*} $, the semi-supervised learning problem takes the form of inference for a Boltzmann machine with hidden units $\observationSample^{*}$ corresponding to the set of unknown labels. The quantity of interest is:

\begin{equation}\label{eq:conditionalSemi}
Pr( \observationSample^{*} | \observationSample)[ \sample^{*} \cup \sample ] 
\end{equation}

This obeys the relation:

\begin{equation}
Pr( \observationSample^{*}  | \observationSample ) [ \sample \cup \sample^{*} ] = \frac{Pr( \observationSample^{*} , \observationSample )[ \sample \cup \sample^{*} ] }{Pr( \observationSample )  [ \sample \cup \sample^{*} ] } \propto Pr( \observationSample^{*} , \observationSample ) [ \sample \cup \sample^{*} ]
\end{equation}

Here proportionality $\propto$ means the conditional distribution for $Pr( \observationSample^{*}  | \observationSample ) [ \sample \cup \sample^{*} ]$ will have the same functional form for $\observationSample^{*}$ as $Pr( \observationSample^{*} , \observationSample ) [ \sample \cup \sample^{*} ]$ up to a multiplicative constant. This constant of proportionality will not affect the MAP assignment for $\observationSample^{*}$ which we will target for semi-supervised learning.

We define an `energy' $E(\observationSample^{*})$ that obeys:

\begin{equation}
Pr( \observationSample^{*} , \observationSample ) [ \sample \cup \sample^{*} ] = \exp \lbrace -E(\observationSample^{*}) \rbrace 
\end{equation}

The MAP estimate for $\observationSample^{*}$ corresponds to the minimum energy configuration. Following the convention of Boykov et al \citeyearpar{Boykov2001}, we consider the energy to be a sum of three terms- a `data' term, a `smoothing' term,  and a constant term $c$ that does not depend on $\observationSample^{*}$.

\begin{equation}
E(\observationSample^{*}) = E_{\hbox{data}}( \observationSample^{*} ) + E_{\hbox{smooth}}( \observationSample^{*} ) + c
\end{equation} 

To ascertain the first two terms we compare to equation (\ref{eq:explicitMRF}). The terms are given by:

\begin{equation}
-E_{\hbox{data}}( \observationSample^{*} ) = \sum_{\observation^{*}_i}^{|\sample^{*}|} \meanFunction^{(\observation^{*}_i)} + \frac{1}{2}\sum_{\pointIndex=1}^{|\sample^{*}|} \covarianceFunction^{(y^{*}_i)}(\mathbf{0}) + \sum_{\pointIndex=1}^{|\sample|} \sum_{\pointIndexB=1}^{|\sample^{*}|}\delta(\observation_\pointIndex,\observation^{*}_\pointIndexB)\covarianceFunction^{(\observation_\pointIndex)}(x_\pointIndex,x^{*}_\pointIndexB)
\end{equation}

\noindent and:

\begin{equation}\label{eq:smoothingEnergy}
-E_{\hbox{smooth}}( \observationSample^{*} ) = \sum_{ \pointIndex< \pointIndexB = 1} ^{ |\sample^{*}| } \delta(y^{*}_\pointIndex, y^{*}_\pointIndexB) \covarianceFunction^{(y^{*}_\pointIndex)}( x_\pointIndex^{*} - x_\pointIndexB^{*} )
\end{equation}

Note that we no longer have a factor of $1/2$ in equation (\ref{eq:smoothingEnergy}) because we now sum only over $\pointIndex< \pointIndexB$. We write it this way for consistency with Boykov et al \citeyearpar{Boykov2001}.

Recall that all kernel functions we consider are non-negative. Under these conditions, as we will show in the next section \ref{section:submodularity}, it is possible to exploit graph min-cut methods. In the case of binary classification this will give the exact MAP state in low order polynomial time \citep{Kolmogorov2004}. For more than two states we can no longer guarantee to find the MAP estimate in polynomial time. However we can use the expansion algorithm \citep{Boykov2001} to find a strong local optimum. 

As has already been mentioned, semi-supervised learning using general Markov random fields has been considered before in the machine learning community \citep{Zhang2001,Blum2001}. The latter of these is the closest to the current work since it uses graph min-cut algorithms. The comparison here is informative. Our work derives the model from a log Gaussian Cox process which is not true of any algorithm in the existing literature. The derivation from a log Gaussian Cox process places additional requirements on the energy functions whilst (as we shall prove) still maintaining the submodularity necessary for efficient inference. As well as being interesting from a Bayesian nonparametric perspective, these extra requirements seem to answer the difficulty that Blum and Chawla describe in motivating and choosing the form of energy function for their algorithm, when viewed as an MRF. One potential issue with using the global MAP assignment is that it is the optimal decision under a global $0$-$1$ loss rather than the Hamming loss. Blum et al discuss these issues \citeyearpar{Blum2004} and advocate an ensemble based on randomizing the potential functions. The empirical improvements they observe can perhaps in hindsight be related to recent work on approximately sampling MRFs in the `strong data\slash strong coupling' regime using randomized MAP methods \citep{Hazan2013}. We expect such considerations are relevant here but they are beyond the scope of the current paper.


\subsection{Pairwise graph representability and submodularity of Markov random field}{\label{section:submodularity}}

In this section we describe new results on the pairwise graph representability and submodularity of the model. This is useful because it means it is possible to use low order polynomial time graph min-cut inference. 

Kolmogorov and Zabih \citeyearpar{Kolmogorov2004} define the term \emph{graph-representable} for a binary energy minimization problem that can be represented in such a way that it is possible to find its minimum using graph min-cut algorithms. Min-cut algorithms find the exact minimum energy configuration of the binary problem.  The same work contains the following result:

\begin{kAndZ}[Kolmogorov and Zabih \citeyearpar{Kolmogorov2004}] A real valued energy function $E(\observationSample^{*} )$ with binary inputs $\observationSample^{*}  \in \lbrace 0, 1 \rbrace ^{\nTrainingPoints} $ and:

\begin{equation}
E(\observationSample^{*} ) = \sum_{\pointIndex} E_{\pointIndex}(y^*_\pointIndex) + \sum_{\pointIndex<\pointIndexB=1}^{\nTrainingPoints} E_{(\pointIndex,\pointIndexB)}(y^*_\pointIndex,y^*_\pointIndexB)
\end{equation}

\noindent is graph-representable if and only if: 

\begin{equation}
E_{(\pointIndex,\pointIndexB)}(0,0) + E_{(\pointIndex,\pointIndexB)}(1,1) \leq E_{(\pointIndex,\pointIndexB)}(0,1) + E_{(\pointIndex,\pointIndexB)}(1,0) 
\end{equation}

\noindent for all pairs $(\pointIndex,\pointIndexB)$.

\end{kAndZ}

For a proof of this result and further discussion see the original work. For more than two classes the expansion algorithm breaks the problem down repeatedly into a binary problem and then applies a graph min-cut algorithm. Although the expansion algorithm is not guaranteed to find the exact minimum energy state it finds a strong local optimum. For more discussion of the quality of this approximation see the thesis of Veksler \citeyearpar{Veksler1999}. The condition for the expansion algorithm to be valid is that the energies satisfy \citep{Kolmogorov2004}:

\begin{equation}{\label{eq:submodularInequality}}
E_{\pointIndex,\pointIndexB}(a,a) + E_{\pointIndex,\pointIndexB}(b,c) \leq E_{\pointIndex,\pointIndexB}(a,c) + E_{\pointIndex,\pointIndexB}(b,a)
\end{equation}

\noindent for any three classes $a,b,c$ and sites $\pointIndex,\pointIndexB$. We refer to this condition as \emph{pairwise graph-representability}. It can be seen that pairwise graph-representablity corresponds to graph-representablity when we are considering the binary case only. We now give a new result.

\begin{mAndG}The energy minimization problem corresponding to the maximum a posteriori state of the log Gaussian Cox process semi-supervised learning problem is pairwise graph representable.
\end{mAndG}

To prove this we need to show that condition (\ref{eq:submodularInequality}) holds for the energy functions in our problem. It suffices only to consider the pairwise terms given in equation (\ref{eq:smoothingEnergy}). That is we take:

\begin{equation}
E_{\pointIndex,\pointIndexB}(y^{*}_\pointIndex, y^{*}_\pointIndexB) = E_{\pointIndex,\pointIndexB}(y^{*}_\pointIndex, y^{*}_\pointIndexB)_{\hbox{smooth}} = - \delta(y^{*}_\pointIndex, y^{*}_\pointIndexB) \covarianceFunction^{(y^{*}_\pointIndex)}( x_\pointIndex^{*} - x_\pointIndexB^{*} )
\end{equation}

We simply need to exhaust each relevant possibility for the relationship between the labels. That is to say we know that:

\begin{equation}
(a=b=c)\veebar( (a=b) \land (b\neq c) )\veebar( (a\neq b) \land (b=c) ) \veebar( (a\neq b) \land (b \neq c ) )
\end{equation}

\noindent where $\veebar$ denotes exclusive logical disjunction. We then substitute each case into the inequality to verify it is correct. We will need to use the condition that all covariance functions are assumed to be non-negative.

Pairwise graph representable problems are a subset of pairwise submodular problems \citep{Kolmogorov2004}. Hence a corollary of our theorem is that the problem is pairwise submodular. More general submodular solvers are not currently generally as fast as graph min-cut solvers where the latter are applicable since the former cannot exploit the extra specific structure present in the problem.   

\subsection{Parameters}

The effect of multiplying mean and covariance functions by some common value $\gamma>1$ is analogous to changing the temperature in a Markov random field. It will increase the confidence of predictions in equation (\ref{eq:predictiveDistribution}) but it will not change the order of the probabilities and hence not change the most likely category.

Under the stationarity assumption, we have a constant mean function for each category. If we consider the effect on equation (\ref{eq:predictiveDistribution}) of changing the means, we can see that it allows us to model a prior bias between categories, independent of the training data.

\section{Experiments}

To perform experiments we created software in MATLAB, building on the GPML toolbox \citep{Rasmussen2010}. For the graph min-cut algorithms we used the code that accompanies the original papers \citep{Boykov2001,Boykov2004,Kolmogorov2004}. Comparisons to the support vector machine (SVM) where performed using SVM-light \citep{Joachims1999}. The harmonic functions algorithm we used for comparison in the semi-supervised experiments was a slight modification of the code accompanying the original paper \citep{Zhu2003}.

\subsection{Experiments on demonstration data}

In order to demonstrate the LGCP classifier we first show performance on an illustrative synthetic dataset. Figure (\ref{fig:concentricCircles}) (left) shows a generated dataset of two concentric circles. The outer training data is clearly not in a convex set, but this is not an issue for the method. Figure (\ref{fig:concentricCircles}) (centre) shows the result of performing prediction on a notional test point using equation (\ref{eq:predictiveDistribution}). The mean functions were taken to be zero and an isotropic squared exponential kernel with signal standard deviation $0.5$ and length scale $1$ was used. It can clearly be seen that the model gives sensible predictions on this test data. Figure (\ref{fig:concentricCircles}) (right) shows a semi-supervised version of the same experiment with the same fixed parameters for the kernel. Using the graph min-cut methods described in section \ref{section:SSL} the labels are recovered with only one error, made on a point that is close to the margin between classes. We performed the analogous experiment for three concentric circles, again using the graph min-cut algorithm for the semi-supervised learning, and obtained similar good results.

\begin{figure}[h!]
\vskip 0.2in
\begin{center}
\centerline{\includegraphics[width=0.35\columnwidth]{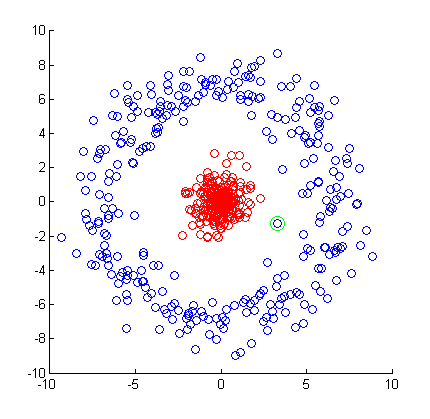}\includegraphics[width=0.35\columnwidth]{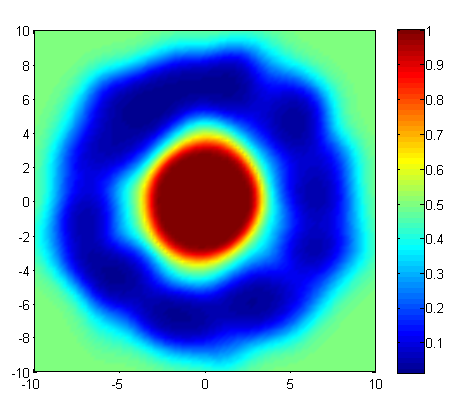}\includegraphics[width=0.35\columnwidth]{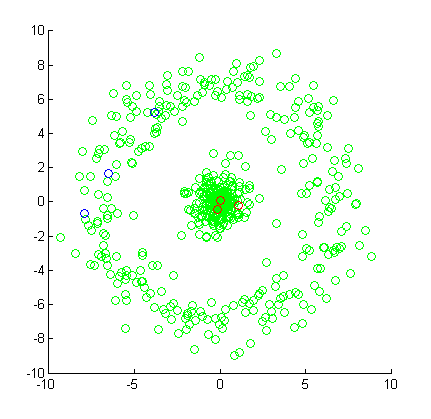}}
\caption{Left: Generated data with two concentric circles used for demonstrating LGCP. Centre: Predictive probability for a notional test point in the 2D plane for the same data. Right: A semi-supervised version of the same experiment. The labelled data are shown in red and blue and the unlabelled data are shown in green. The labels are recovered with only one error circled in green in the left panel.}
\label{fig:concentricCircles}
\end{center}
\vskip -0.2in
\end{figure}

\subsection{Experiments on supervised classification datasets}\label{section:supervisedResults}

Experiments were performed on six commonly used classification datasets. The Synth dataset, and the Diabetes dataset come from Ripley \citep{Ripley1996} \footnote{Data available from http://www.stats.ox.ac.uk/pub/PRNN/}. The Banana dataset originally comes from R{\"a}tsch \footnote{Data available from http://mldata.org/}.They are commonly used for comparison of classification algorithms, see for instance \citep{Naish2008}. The Brain Computer Interface (BCI)  dataset is originally used in the work of Lal et al \citeyearpar{Lal2004}. The covariates correspond to the parameters of a time series model fitted to electroencephalography data taken from a single subject. The categories correspond to the subject imagining a movement either with the left or right hand. The restricted STL (rSTL) dataset is a subset of the full STL image dataset \citep{Coates2011}. The dataset consists of 96$\times$96 RGB images. We randomly selected 500 images each of monkeys and trucks. Both the BCI dataset and rSTL dataset are relatively challenging classification tasks where the number of data points is in some sense small relative to the dimensionality of the data. The MNIST handwritten digit dataset \citep{Lecun1998,MNIST2013} is a very highly investigated classification dataset. The data has a relatively large number of data points, is multiclass, and has relatively high dimensional covariates.

For the Synth, Diabetes, and MNIST datasets we used the standard partition to aid comparison to other papers in the literature. For the Banana dataset, BCI and rSTL we randomly partitioned between train and test.

The experiments were performed using the LGCP classifier with zero mean function. All experiments used the isotropic squared exponential kernel, which was shared between classes. In all cases the single shared length scale was the only free parameter. It was fitted using leave-out-one cross validation with a $0$-$1$ loss. The parameter training was done using all the training points, with the first five smaller datasets and $1000$ randomly chosen points from the MNIST training set. In the latter case this was sufficient for our needs, given that we were only training one parameter. One iteration of cross validation over all $1000$ points, for a given length scale, on this MNIST subset took 17 seconds on a single Intel I3 2.53 GHz core with 4GB of RAM. The prediction on the validation data for MNIST using all 60000 training points and 10000 validation points took 130 seconds on the same computer. 

For the datasets other than MNIST we compared to the support vector machine (SVM) and Gaussian process classification (GPC) both using the same kernel. For GPC we used variational approximate inference, the logistic link function, and zero mean function. To mitigate any issues with local optima we repeated the GPC gradient descent with 10 different random initializations and took the run with the best marginal likelihood. The kernel length scale of the SVM was trained using leave-out-one cross validation whilst the regularization parameter was set using the default heuristic of the software. For the MNIST dataset a wide variety of algorithms have been tested under controlled conditions and the results published. Insofar as we are aware there is no published result for GPC on the MNIST dataset which is surprising because of the significance of such a result. We speculate that this is due to the technical challenges of providing multiclass, sparse, approximate inference on a large high dimensional dataset. 

\begin{table}[H]
\caption{Error \textbackslash \%  for supervised binary datasets}
\label{tb:supervisedResults}
\begin{center}
\begin{small}
\begin{sc}
\begin{tabular}{|l | cc | ccr | }
\hline
{\bf Dataset} & {\bf Train \slash Test } &  {\bf Variables  \slash Classes }  & { \bf LGCP } &  {\bf GPC } &{\bf SVM}  \\
\hline
Synth & 250 \slash 1000 &  2  \slash 2  & 9.1  &  10.1 & 9.4 \\
Diabetes & 200 \slash 332 &  7 \slash  2  & 21.7  &  20.8 &  22.3 \\
Banana & 400 \slash 4900 & 2 \slash 2 & 11.4 &  10.6 &  10.7 \\
rSTL & 500 \slash 500 & 27648 \slash 2 & 20.0 & 18.6 & 12.4  \\
BCI & 200 \slash 200 & 117 \slash 2 & 39.5 & 24.5 & 31.5  \\
\hline
\end{tabular}
\end{sc}
\end{small}
\end{center}
\end{table}

\begin{table}[H]
\vskip -0.2in
\caption{Error \textbackslash \%  for Supervised MNIST Dataset}
\label{tb:results}
\begin{center}
\begin{small}
\begin{sc}
\begin{tabular}{|lc|ccccccr|}
\hline
{\bf Train \slash Test } & {\hspace{-6 pt} \bf Var  \slash Classes } & { \bf \hspace{-3 pt} LGCP } &  {\bf \hspace{-3 pt} 1-NN } &{\bf \hspace{-3 pt} SVM} & {\bf \hspace{-3 pt} FF. net } & { \bf \hspace{-3 pt} Log. reg. } &  {\bf \hspace{-3 pt} RBF } & {\bf \hspace{-3 pt} Conv. net } \\
\hline
60K\slash 10K & 784 \slash 10 & 3.11  &  3.09  & 1.4 & 4.5 & 12.0  &  3.6  & 0.23 \\
\hline
\end{tabular}
\end{sc}
\end{small}
\end{center}
\end{table}
\vskip -0.1 in

The results obtained for the datasets excluding MNIST are shown in table \ref{tb:supervisedResults}. The results for LGCP on the Synth, Diabetes and Banana datasets are comparable with the SVM and GPC. This is the case despite both of these two alternative algorithms having more free parameters than the single parameter LGCP. The LGCP classifier lags the SVM and GPC on the two higher dimensional datasets BCI and rSTL although neither of these alternatives is better outright.

The results for LGCP on the MNIST dataset, shown in table \ref{tb:results}, are similar to those published for $K$-nearest neighbours (KNN) with a Euclidean $L2$-norm. Given the links to classical nonparametric models described in section \ref{section:classical} this is perhaps not surprising. For sufficiently short length scales the sum over training points in equation (\ref{eq:activationFunction}) is dominated by the term with the smallest Euclidean distance, i.e the nearest neighbour. Conditions under which kernel sum based algorithms will be similar to KNN are discussed in the literature \citep{Bengio2005}. If the data lie near a low dimensional manifold the squared exponential kernel will only depend on the dimensionality of this manifold \citep{Tenenbaum2000} which is also of relevance to semi-supervised learning.

The method is invariant to joint permutation of the covariates. Despite its simplicity it outperforms some other more complex methods \citep{Lecun1998,MNIST2013} such as early experiments with the feed forward neural network (4.5\%). The SVM with the same kernel performs well on this dataset with an error of 1.4\%. The best current performance is achieved by deep convolutional neural networks, which can achieve very low error rates around 0.23\% \citep{Ciresan2012}. These methods are not permutation invariant, incorporate strong prior knowledge, and require a good deal of training.

The LGCP algorithm has scope for generalization since there is a wide variety of mean functions and covariance functions that would encode more complex inductive biases, particularly if as we suspect, the assumption of stationarity may be relaxed. Since this would lead to a larger number of parameters it would be necessary to leverage the existing literature on approximate learning in Markov random fields \citep{Koller2009}.

\subsection{Semi-supervised experiments}\label{section:sslExperiments}

We also performed experiments to compare SVM, GPC, LGCP, semi-supervised LGCP (SLGCP ) and the semi-supervised Harmonic functions algorithm of Zhu et al \citeyearpar{Zhu2003}.

To compare the algorithms we used three different datasets. The double helix dataset is a synthetic dataset inspired by the paper of Zhu et al \citeyearpar{Zhu2003}. An advantage of this dataset is that it is a non-trivial classification task where it is still possible to visualize the geometric advantage that semi-supervised methods have over supervised methods. The parameters were chosen in order to give a reasonable separation between the various methods. The dataset is shown in figure (\ref{fig:doubleHelix3D}). For the next dataset we took the digits 0 and 4 from the MNIST training set and randomly sampled 500 examples from each class. We call the resulting dataset reduced MNIST or rMNIST. The third dataset consisted in data from the Oil pipe dataset. This dataset was first considered for machine learning in the work of Bishop and James \citeyearpar{Bishop1993}. The covariates consist in attenuation measurements of gamma rays passed through an oil pipe. In our case we randomly select 500 examples each corresponding to homogeneous and annular flow and the task is to distinguish between these two classes.

\begin{figure}[h]
\begin{center}
\centerline{\includegraphics[width=0.8\columnwidth]{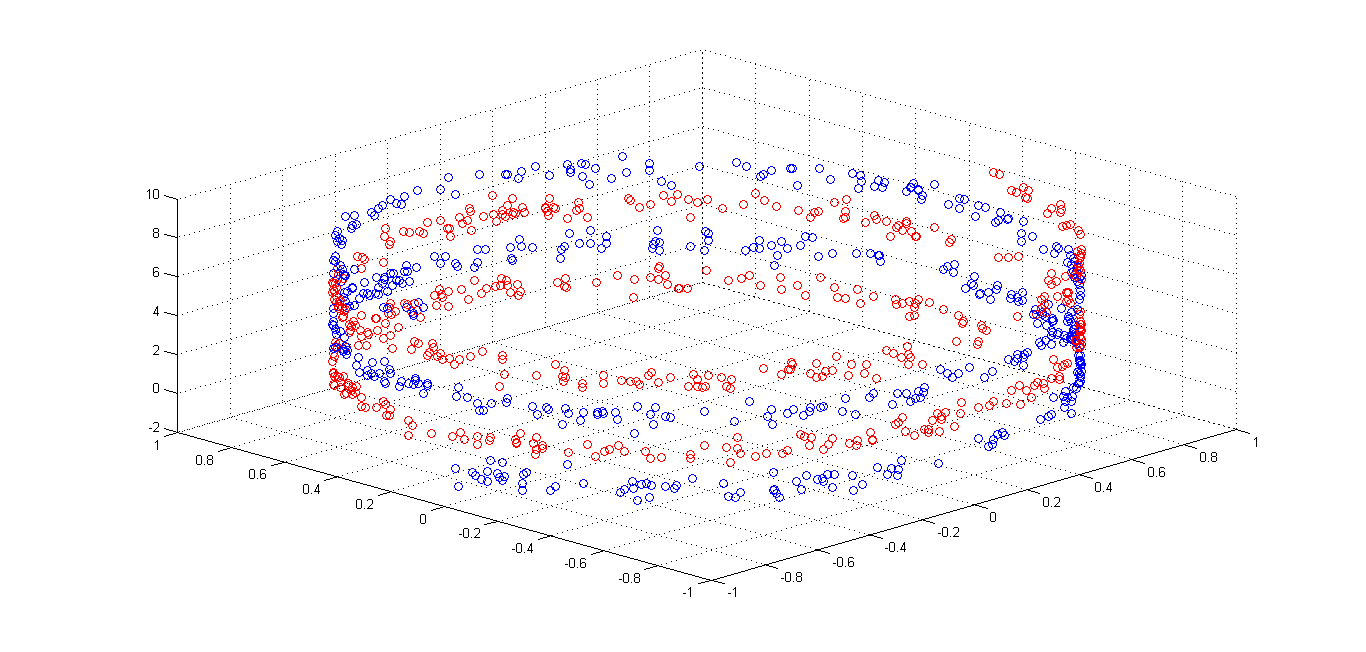}}
\caption{ The double helix dataset. The red and blue points show the two different classes to be classified.}
\label{fig:doubleHelix3D}
\end{center}
\vskip -0.2in
\end{figure}

For each dataset we varied the number of labelled points and also performed 10 randomized training / test partitions. Before comparing the algorithms we ruled out the existence of a trivial semi-supervised solution in terms of a bipartite nearest neighbour graph on the Euclidean distance between covariates.

All supervised algorithms were set up as in section \ref{section:supervisedResults}. The Harmonic functions algorithm was given an isotropic squared exponential weight function. The single length scale hyperparameter was trained using cross validation. The SLGCP again had one free parameter, the common length scale of the covariance function and this was fixed using 10 fold cross validation on the $0$-$1$ loss. The Harmonic functions algorithm and SLGCP were trained `transductively' ie. they had access to the test points as unlabelled data during training. With 10 labelled points, at the optimal length scale, the 10 folds of cross validation for SLGCP took 1 minute and 25 seconds on the same computer described in section \ref{section:supervisedResults}.

\begin{figure}[h]
\begin{center}
\centerline{\includegraphics[width=0.8\columnwidth]{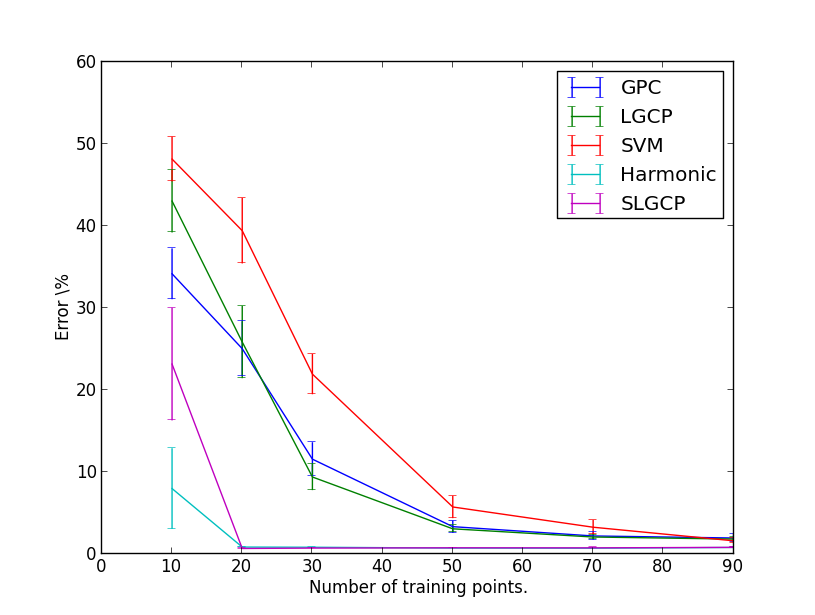}}
\caption{ Test error for the double helix data set as a function of number of training points. The error bars are the standard error of the mean with averaging over 10 random partitions.}
\label{fig:sslDoubleHelix}
\end{center}
\vskip -0.2in
\end{figure} 

\begin{figure}[h]
\begin{center}
\vskip -0.2in
\centerline{\includegraphics[width=0.8\columnwidth]{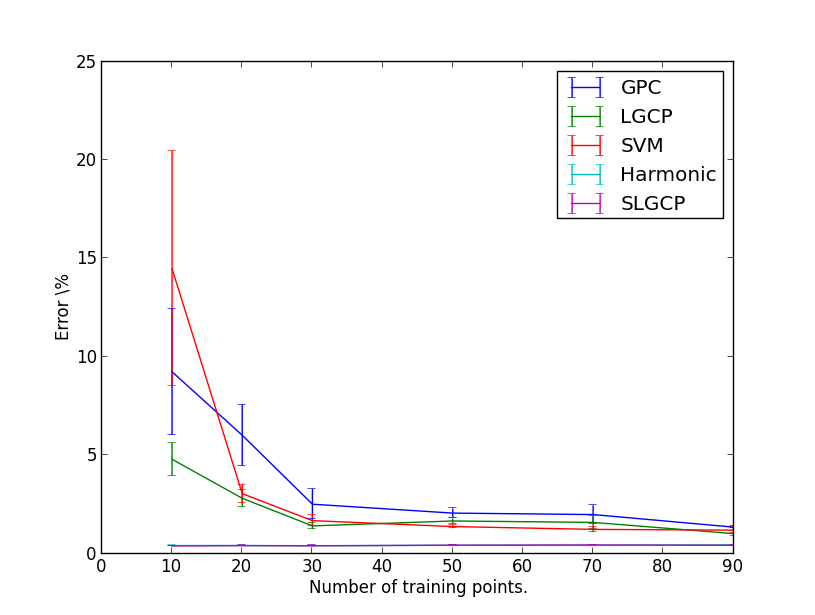}}
\caption{ Test error for the rMNIST data set as a function of number of training points. The error bars are the standard error of the mean with averaging over 10 random partitions. The harmonic functions algorithm is indistinguishable from SLGCP (see text). }
\label{fig:sslExperiment}
\end{center}
\vskip -0.2in
\end{figure} 

\begin{figure}[h]
\begin{center}
\vskip -0.2in
\centerline{\includegraphics[width=0.8\columnwidth]{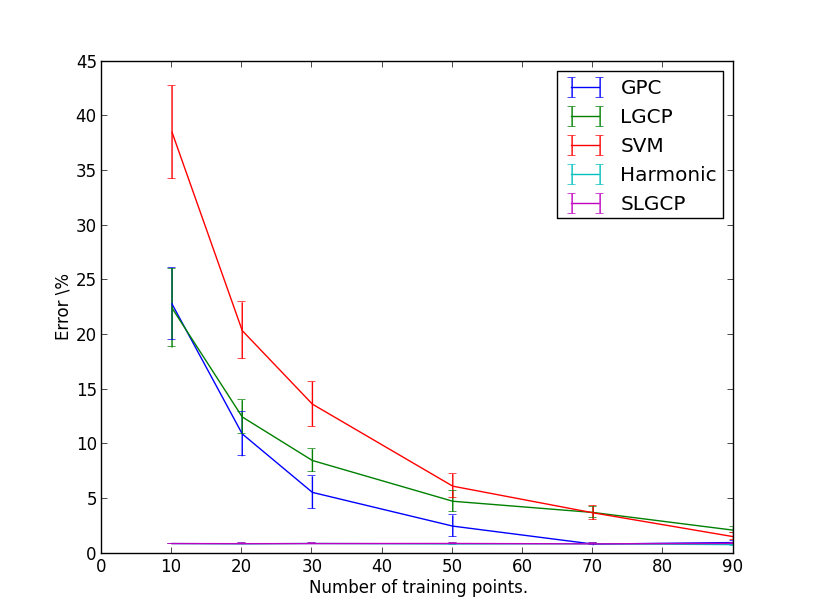}}
\caption{ Test error for the oil pipe data set as a function of number of training points. The error bars are the standard error of the mean with averaging over 10 random partitions. The harmonic functions algorithm is indistinguishable from SLGCP (see text). }
\label{fig:sslOil}
\end{center}
\vskip -0.2in
\end{figure} 

Figures (\ref{fig:sslDoubleHelix}), (\ref{fig:sslExperiment}) and (\ref{fig:sslOil})  show the results of the experiment. As we would expect the semi-supervised SLGCP generally outperforms its supervised counterpart LGCP and all the other supervised methods, often by a substantial margin. In the majority of cases SLGCP and the Harmonic functions algorithms perform similarly well the only exception being for the smallest number of training points on the double helix dataset where the Harmonic functions algorithm has a lower error rate. A common feature of the datasets is that for each class the covariates lie near a low dimensional manifold in the higher dimensional space, separated by a complex decision surface and the unlabelled data can make a large difference in the quality of prediction by helping to identify these manifolds. Our results show the SLGCP exploiting this advantage.

\section{Conclusions}

In this paper we extended the LGCP model to semi-supervised tasks and gave the first practical demonstration of the LGCP model on supervised and semi-supervised tasks. The linear time supervised prediction meant that we were able to demonstrate relatively large supervised datasets. Despite its simplicity, there are domains where performance is comparable to well established methods. We gave the first full description of the Markov random field showing that it was pairwise submodular- a necessary property to use graph min-cut methods. We also described intriguing novel connections to classical nonparametric methods.

In terms of further work a promising avenue is to incorporate more complex inductive biases into the mean and covariance functions, exploiting literature on Markov random fields \citep{Koller2009} for parameter learning. For the semi-supervised learning method, as mentioned in section \ref{section:SSL} there is scope to combine the existing literature on randomized MAP with these methods \citep{Blum2004,Hazan2013}. In Bayesian nonparametric terms this method is an example of the largely untapped potential for combining the rich theory of spatial point processes with the important application domain of classification and we believe this is a promising area for further research.

\section{Acknowledgements}

We would like to acknowledge the support of EPSRC grant EP/I036575/1 and a Google Focussed Research Award. We would like to thank Peter McCullagh for comments that improved the presentation and clarity of the paper.

\appendix

\section*{Appendix: Some comments on the comparison between LGCP and the permanental Cox process classifier}\label{section:permanental}

In this section we compare classification based on the permanental Cox process \citep{McCullagh2006b} to the LGCP classifier. The comparison has been made before in the existing literature theoretically \citep{McCullagh2006,McCullagh2008} and the permanental process classifier has been studied experimentally \citep{McCullagh2012}. Here we provide some summary comments on this existing literature and discuss practical insights from our recent work on the LGCP, with a particular bias towards a machine learning perspective. 

For the permanental Cox process both the product density and the Janossy density are available in closed form \citep{McCullagh2006b}. We define a new function $\hat{C} = 2C$. The product density is given by:

\begin{equation}
m_{[\nProduct]}\densityArgument = \operatorname{Per}_{\alpha}(\hat{\covarianceFunction}[\mathbf{x}]).
\end{equation}

Here $\operatorname{Per}_{\alpha}(B)$ is the alpha permanent of a matrix $B$. The local Janossy density with respect to Lebesgue measure on $\densitySpace$ and bounded Borel set $\testSet$ is given by

\begin{equation}
\janossyDensity\localDensityArgument = \exp\lbrace-\alpha D \rbrace\operatorname{Per}_{\alpha}(\tilde{\covarianceFunction}[\mathbf{x}])
\end{equation}

\noindent where $\tilde{\covarianceFunction}$ is defined in terms of the eigenvalues $\lbrace \lambda_i \rbrace$ and eigenfunctions $ \lbrace \mathbf{e}_i \rbrace $ of $\hat{C}$ with respect to Lebesgue measure over $\testSet$ using the relation:

\begin{equation}
\tilde{\covarianceFunction}(x,x') = \sum_{i=1}^{\infty} \frac{\lambda_i}{(1+\lambda_i)} \mathbf{e(x)}\mathbf{e(x')}
\end{equation}

\noindent and $D = \sum_{i=0}^{\infty} \log(1+\lambda_i)$. We may in fact obtain closed form predictive equations for classification using the product density \citep{McCullagh2008} or using the Janossy density \citep{McCullagh2012}. By contrast a closed form is only available for the product density in the LGCP case \citep{Moller1998}.

Unlike the log-Gaussian Cox process, the family of permanental Cox processes is closed under superposition which is another appealing theoretical property. Further, McCullagh and Yang \citeyearpar{McCullagh2006} are able to extend their classification model to an unbounded number of classes, effectively deriving a kernel generalization of the Chinese restaurant process. It seems difficult to imagine a similar extension for log-Gaussian Cox processes. 

The catch however lies in the difference between `closed form' and tractable. The alpha permanent of a matrix is $\numberP$-Hard to compute \citep{Valiant1979} and although approximating the ratio of two related alpha permanents seems to be an easier problem \citep{Kou2009}, even approximating this ratio for prediction currently requires $O(\nTrainingPoints^3)$ \citep{McCullagh2012} for the four-cycle approximation as compared to the $O(\nTrainingPoints)$ for the LGCP. This means that the log Gaussian variant will be of relatively more interest when scalability is a factor. Further, the connection between the log Gaussian variant and the Potts model \citep{McCullagh2008} allows progress to be made on the semi-supervised learning problem and the supervised case has interesting links to classical non-parametric estimators. Thus we believe that both variants merit continuing investigation.

\bibliography{../bib/bibliography}
\bibliographystyle{apalike} 

\end{document}